\documentclass{article}

\usepackage{microtype}
\usepackage{graphicx}
\usepackage{subfigure}
\usepackage{booktabs} 
\usepackage{graphicx}
\usepackage{hyperref,color,soul,booktabs}
\setulcolor{blue}
\usepackage{wrapfig} 
\usepackage{tikz}
\usepackage{caption}
\usepackage{multirow}
\DeclareFontFamily{OT1}{pzc}{}
\DeclareFontShape{OT1}{pzc}{m}{it}{<-> s * [1.190] pzcmi7t}{}
\DeclareMathAlphabet{\mathpzc}{OT1}{pzc}{m}{it}
\DeclareMathAlphabet{\mathcal}{OMS}{cmsy}{m}{n}

\def\hlinewd#1{%
\noalign{\ifnum0=`}\fi\hrule \@height #1 %
\futurelet\reserved@a\@xhline} 

\makeatother

\makeindex
\makeglossary
\makeatletter

\newcommand{\Rmnum}[1]{\expandafter\@slowromancap\romannumeral #1@}
\makeatother
\makeatletter
\newcommand*{\rom}[1]{\expandafter\@slowromancap\romannumeral #1@}
\makeatother

\usepackage{hyperref}

\usepackage[accepted]{icml2020}

\icmltitlerunning{}

\begin{document}

\twocolumn[
\icmltitle{Hierarchical Dependency Constrained Tree Augmented Na\"{i}ve Bayes Classifiers for Hierarchical Feature Spaces}




\begin{icmlauthorlist}
\icmlauthor{Cen Wan}{to}
\icmlauthor{Alex A. Freitas}{goo}
\end{icmlauthorlist}

\icmlaffiliation{to}{Birkbeck, University of London, London, United Kingdom}
\icmlaffiliation{goo}{University of Kent, Canterbury, United Kingdom}

\icmlcorrespondingauthor{Cen Wan}{cen.wan@bbk.ac.uk}


\vskip 0.3in
]



\printAffiliationsAndNotice{} 

\begin{abstract}
The Tree Augmented Na\"{i}ve Bayes (TAN) classifier is a type of probabilistic graphical model that constructs a single-parent dependency tree to estimate the distribution of the data. In this work, we propose two novel Hierarchical dependency-based Tree Augmented Na\"{i}ve Bayes algorithms, i.e. Hie--TAN and Hie--TAN--Lite. Both methods exploit the pre-defined parent-child (generalisation-specialisation) relationships between features as a type of constraint to learn the tree representation of dependencies among features, whilst the latter further eliminates the hierarchical redundancy during the classifier learning stage. The experimental results showed that Hie--TAN successfully obtained better predictive performance than several other hierarchical dependency constrained classification algorithms, and its predictive performance was further improved by eliminating the hierarchical redundancy, as suggested by the higher accuracy obtained by Hie--TAN--Lite.
\end{abstract}

\section{Introduction}
\indent This work proposes two new Hierarchical dependency constrained Tree Augmented Na\"{i}ve Bayes classifiers (called Hie--TAN and Hie--TAN--Lite) for the classification task of machine learning. Both methods consider the pre-defined hierarchical dependencies between features within a tree or a directed acyclic graph (DAG) as a type of constraint to learn the tree-based representation of feature dependencies, whilst the latter further exploits the hierarchical dependency information to eliminate the hierarchical redundancy between features.

The pre-defined hierarchical dependency information (i.e. ancestor-descendant relationships) is informative and can be exploited for different machine learning tasks. Actually, a series of feature selection methods \cite{Wan2014,SDM18,Pablo2020,LASSOFS2,LASSOFS3,SHSEL} have been proposed to reduce the dataset's dimensionality by exploiting the hierarchical feature dependencies. Those methods successfully obtained in general better predictive performance than other feature selection methods that do not exploit hierarchical feature dependencies. In addition, the pre-defined hierarchical dependency information was also exploited for training a regression model~\cite{GraphLASSO1,GraphLASSO2}, and for constructing Bayesian graphical models~\cite{WanACMBCB2015,WanICML,PGM2020}. However, none of the above studies has exploited the hierarchical dependency information to propose new Tree Augmented Na\"{i}ve Bayes (TAN) methods, as proposed in this work. This is an interesting direction, because TAN can obtain good predictive accuracy whilst requiring less computational cost than other Bayesian network classifiers for learning the graphical structure.

In this work the proposed methods are used to analyse bioinformatics datasets of ageing-related genes, where the features (attributes) are derived from the well-known Gene Ontology (GO) database, where each GO term essentially indicates a type of biological function or property for a given gene~\cite{GO2000}. In each of the datasets used in our experiments, the set of all features (i.e. GO terms) is structured as a DAG by using a type of ``is-a'' relationship~\cite{Wan2014}. This type of relationship implies that GO terms at higher levels of the GO hierarchy indicate more generic definitions of gene functions than GO terms at lower levels of the GO hierarchy~\cite{GO2000}. Each instance in our datasets represents a gene, and each gene is classified as pro-longevity or anti-longevity, depending on its effect on the lifespan of an organism~\cite{DeMagalhaes2009}. However, the proposed algorithm is generic enough to be applied to any type of hierarchically structured features. 

The remainder of this paper is organised as follows. Section 2 briefly reviews the background about the conventional TAN classifier, the rules of dependency propagation and the definition of hierarchical redundancy. Section 3 introduces the proposed Hie--TAN and Hie--TAN--Lite methods. Section 4 presents the experimental methodology and results. Finally, Section 5 presents conclusions and future research directions.

\begin{figure*}[!t]
  \centering
  \includegraphics[width=1.0\linewidth]{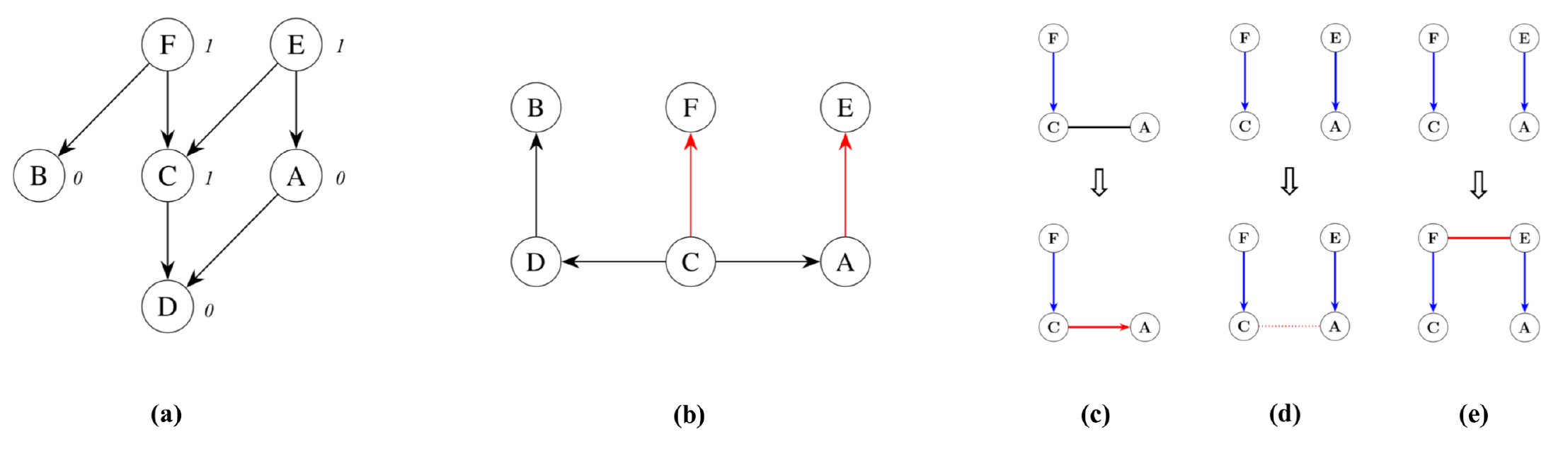}
\caption{\textbf{(a)} A symbolic representation of the GO hierarchy by using 6 arbitrary letters: A, B, C, D, E and F; \textbf{(b)} A network topology learned by the conventional TAN. \textbf{(c-e)} The examples of the dependency propagation process.}
\end{figure*}

\section{Background}
\subsection{Conventional Tree Augmented Na\"{i}ve Bayes}
Tree Augmented Na\"{i}ve Bayes (TAN) is a type of semi-na\"{i}ve Bayes classification algorithm. It avoids the feature independence assumption made by the well-known na\"{i}ve Bayes method, by allowing each feature to have at most one parent feature (i.e. a feature depends on at most one other feature). In addition, the class variable is a parent of all features, as usual in Bayesian network classifiers. TAN firstly generates a ranking of all possible dependencies between pairs of features based on the values of conditional mutual information (CMI), as shown in Equation 1, \noindent where $X_i$ and $X_j$ are predictor features, $Y$ is the class attribute, $x_i, x_j, y$ are the values of the corresponding features and the class attribute, $P(x_i, x_j, y)$ denotes the joint probability of $x_i, x_j, y$; $P(x_i, x_j \mid y)$ denotes the joint probability of feature values $x_i$ and $x_j$ given class value $y$; and $P(x_i \mid y)$ denotes the conditional probability of feature value $x_i$ given class value $y$. Each pair of features ``$x_i, x_j$'' is taken into account as a group, then the mutual information for each pair of features given the class attribute is computed~\cite{Friedman1997}.  Then it builds a maximum spanning tree, which contains the maximum values of conditional mutual information for all edges. Finally, TAN randomly selects a root feature as the starting point to progressively assign the dependencies for all other features in the tree. 
\begin{equation}
\resizebox{.45 \textwidth}{!} 
{
CMI$(X_i;X_j\mid Y)=\sum\limits_{x_i,x_j,y}P(x_i, x_j, y)$log$\frac{P(x_i,x_j\mid y)}{P(x_i\mid y)P(x_j\mid y)}$
}
\end{equation}

However, this approach for dependency assignment usually ignores the pre-defined hierarchical dependencies among features. Figure 1(a) shows an example DAG, where the set of six features is hierarchically structured, e.g. feature F is the parent of features B and C, which is the parent of feature D and also the child of feature E.  Figure 1(b) shows an example network structure learned by the conventional TAN by choosing feature C as the root and then assigning the dependencies between all other features included in Figure 1(a). Note that this learned network does not encode some direct dependencies that occur in the original DAG of Figure 1(a), e.g. the direct dependency between F and B; and conversely, the learned network includes some direct dependencies that do not occur in the DAG of Figure 1(a), e.g. the direct dependency between D and B. In addition, the direction of the A -- E and C -- F edges (shown in red) in Figure 1(b) is the opposite of the direction of these edges in Figure 1(a).

\subsection{Dependency propagation based on existing pre-defined dependencies and the single-parent constraint}
We introduce a dependency propagation rule by considering the single-parent constraint of TAN. As shown in Figures 1(a) and 1(b), given a set of features \{A,...,F\}, and a pre-defined feature hierarchy, we can obtain a list of edges representing all possible pairs of features. Based on a pre-defined feature hierarchy, some edges have a direction, such as edge F $\rightarrow$ C, since feature F is the ancestor of feature C; whilst some edges do not have a direction, since there is no pre-defined hierarchical relationship between features in Figure 1(a) -- e.g. edge C -- A in Figure 1(b). But under certain circumstances, undirected edges can be assigned directions by considering other connected directed edges and the single-parent constraint when learning the structure of a TAN classifier.

For a given set of edges $\mathpzc{E}$ that have already been added into the tree $\mathcal{T}$, if exists one undirected edge $e(X_i,$ $X_j)$, where either $X_i$ or $X_j$ has a parent in any other edge in the set $\mathpzc{E}$, the vertex that doesn't have a parent will be assigned as the child of the other vertex, i.e. $\Pi$$(X_j)$ $\gets$ $X_i$, if $\exists$ $\Pi$($X_i$) $\in$ $\mathpzc{E}$; \textit{vice versa}, $\Pi$$(X_i)$ $\gets$ $X_j$, if $\exists$ $\Pi$($X_j$) $\in$ $\mathpzc{E}$. As shown in Figure 1(c), vertex F is the parent of vertex C, which is also connected with vertex A. Therefore, vertex C is assigned as the parent of vertex A. 

However, when both vertices in the undirected edge already have parents in $\mathcal{T}$, that edge cannot be assigned any direction, since such an assignment would violate the single-parent constraint. For instance, there are two directed edges in Figure 1(d), so when considering adding the undirected edge C -- A, assigning the direction C $\rightarrow$ A or A $\rightarrow$ C would violate the single-parent constraint, since vertices C and A already have parent vertices F and E, respectively. Hence, the undirected edge C -- A will not be added into $\mathcal{T}$. 

The third possible scenario is that both vertices in the undirected edge are parents for some other existing vertices in $\mathcal{T}$. In this case, the undirected edge will be added into $\mathcal{T}$, but the direction cannot be assigned at this stage, since any assigned direction would not violate the single-parent constraint, i.e. $\Pi$$(X_i)$ $\gets$ $X_j$ or $\Pi$$(X_j)$ $\gets$ $X_i$ for a given undirected edge $e(X_i,$ $X_j)$. For instance, as shown in Figure 1(e), when considering an undirected edge F -- E, both vertices F and E are parents of vertices C and A, respectively. So the undirected edge F -- E will be added into $\mathcal{T}$, but it will not be assigned a direction, as shown by the red line. Note that, if there still exists any undirected edges in $\mathcal{T}$ after processing all candidate edges, the directions of those edges will be decided randomly.
\subsection{Hierarchical Redundancy between Features}
For any given instance, the concept of hierarchical redundancy between features in that instance can be defined as follows. Given a feature DAG, if any two given features or vertices in the DAG are connected by a hierarchical relationship, whilst both those features have the same value, those two features (vertices in the DAG) are hierarchically redundant. As shown in Figure 1(a), vertex F is hierarchically redundant to vertex C, since F is the parent of C, and both those features have the same value \textit{1}. Analogously, vertex A is hierarchically redundant to vertex D, since vertex D is the child of vertex A, and both of them have the same value \textit{0}. This type of hierarchical redundancy has been well-studied in previous works ~\cite{WanACMBCB2015,WanICML,PGM2020}, which show that the predictive accuracy of Bayesian network classifiers can be improved by removing this type of redundancy between hierarchically related features. In this work, we further exploit the possibility of removing the hierarchical redundancy whilst also considering the pre-defined hierarchical dependencies to learn the network structure.

\section{Proposed Methods}
\subsection{Hierarchical Dependency Constrained Tree Augmented Na\"{i}ve Bayes}

We first propose a new tree-based Bayesian classifier, named Hierarchical dependency constrained Tree Augmented Na\"{i}ve Bayes (Hie--TAN), which exploits the pre-defined hierarchical dependency information between features to learn the Bayesian classifier's network structure. In general, given a mixed set of directed and undirected edges, the proposed method propagates hierarchical dependencies during the maximum spanning tree learning stage, so that the feature dependencies in the learned TAN network will be based on the pre-defined hierarchical dependencies available in the feature DAG. The method allows each feature to have at most one parent feature (i.e. one feature dependency). Hie--TAN's pseudocode is shown in Algorithms 1 and 2.

\begin{figure}[t]
\includegraphics[width=0.48\textwidth]{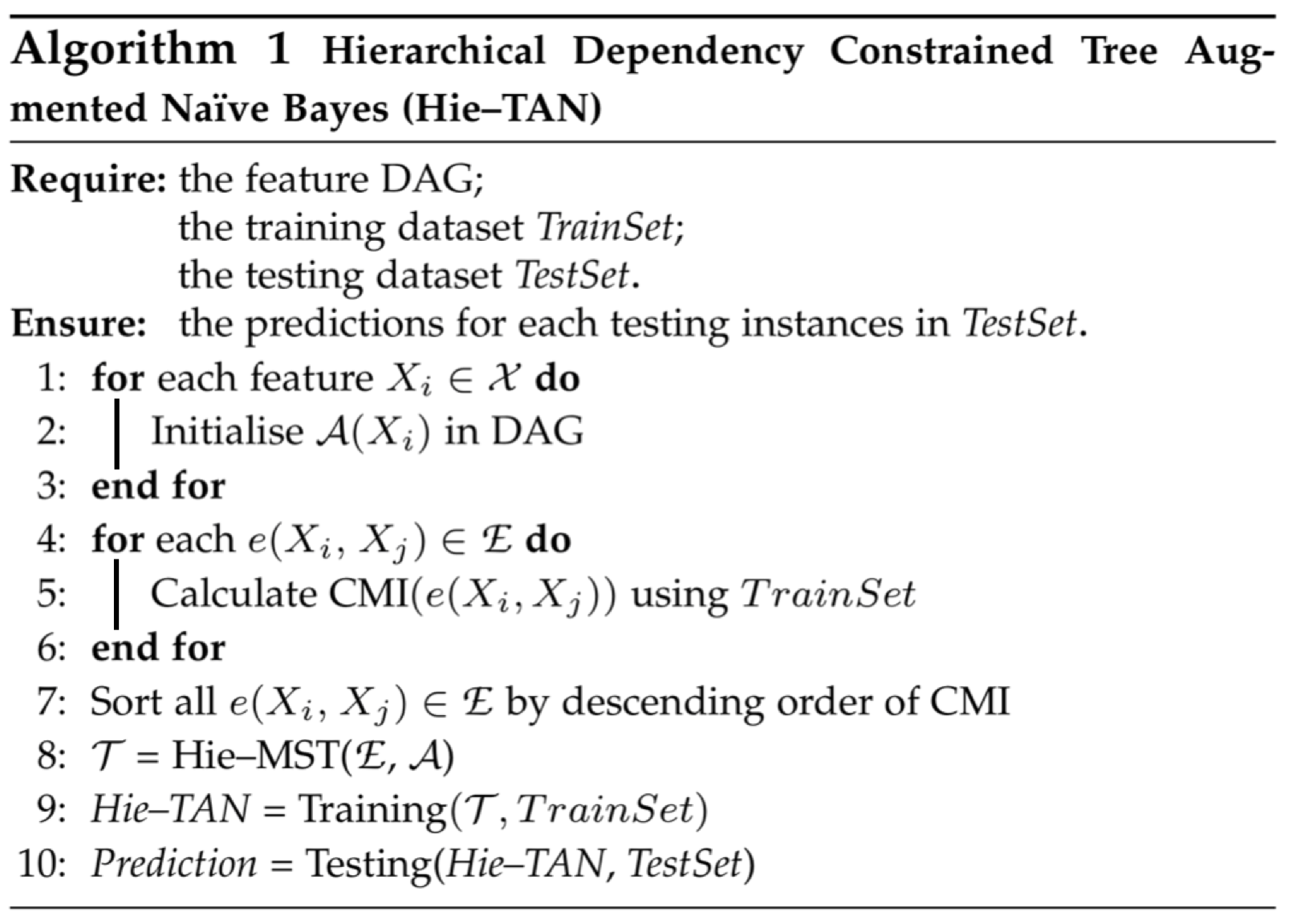}
\end{figure}
In Algorithm 1, in the first part of the Hie--TAN algorithm (lines 1 to 3), Hie--TAN firstly generates the sets of ancestors $\mathcal{A}$ for individual features in $\mathcal{X}$ based on the input feature hierarchy DAG. In addition, the conditional mutual information CMI for all possible edges in $\mathpzc{E}$ is calculated in lines 4 to 6. The values of CMI are used for sorting all edges in $\mathpzc{E}$ in a descending order (line 7). Moreover, those initialised variables are taken as the inputs to the procedure Hie--MST, which learns the tree $\mathcal{T}$ that considers the pre-defined hierarchical dependencies and the single-parent constraint (line 8). Furthermore, the learned tree $\mathcal{T}$ is used to train the Hie--TAN classifier by using the training dataset (line 9). Finally, the trained Hie--TAN classifier is used to predict the class labels of testing instances (line 10).

Algorithm 2 shows the pseudocode of the procedure Hie--MST. It firstly initialises an empty set of directed edges ($\mathcal{D}\mathpzc{E}$) and another empty set of undirected edges ($\mathcal{U}\mathpzc{E}$). Then it processes all individual edges in the sorted $\mathpzc{E}$ in lines 3 to 28. For a given edge $e(X_i,$ $X_j)$, line 4 checks whether adding this edge will lead to a cycle by considering all existing edges in both sets $\mathcal{D}\mathpzc{E}$ and $\mathcal{U}\mathpzc{E}$. If adding that edge $e(X_i,$ $X_j)$ will not lead to a cycle, Hie--MST checks whether vertices $X_i$ and $X_j$ contain a pre-defined hierarchical dependency according to the DAG (line 5). In lines 6 to 9, if vertices $X_i$ and $X_j$ contain a pre-defined hierarchical dependency, Hie--MST will check whether adding that edge leads to the violation of the single-parent constraint (line 6). If not so, edge $e(X_i,$ $X_j)$ will be added into the set $\mathcal{D}\mathpzc{E}$ (line 7). 
  
\begin{figure}[!t]
\includegraphics[width=0.48\textwidth]{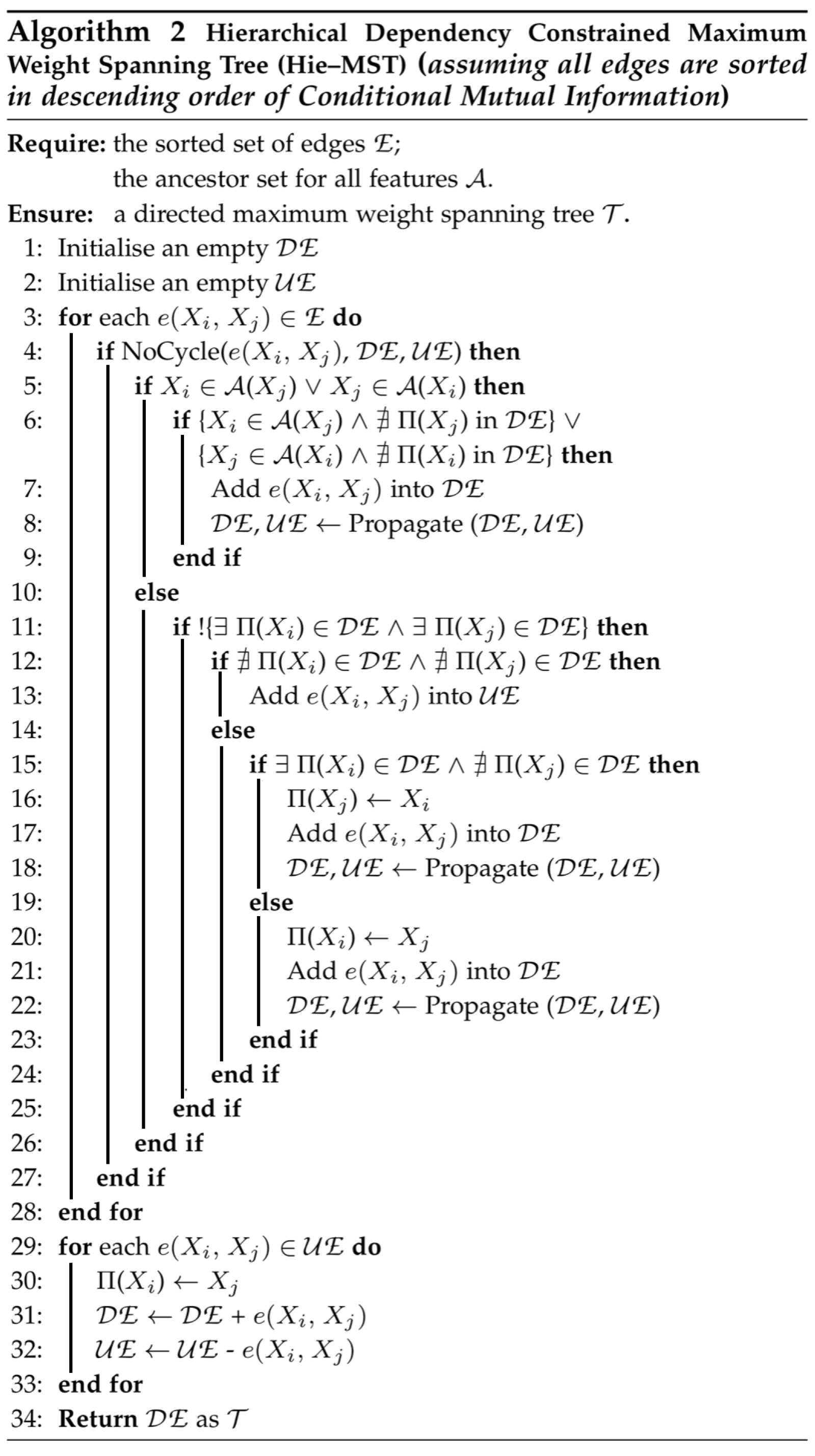}
\end{figure}

\begin{figure}[!t]
\includegraphics[width=0.48\textwidth]{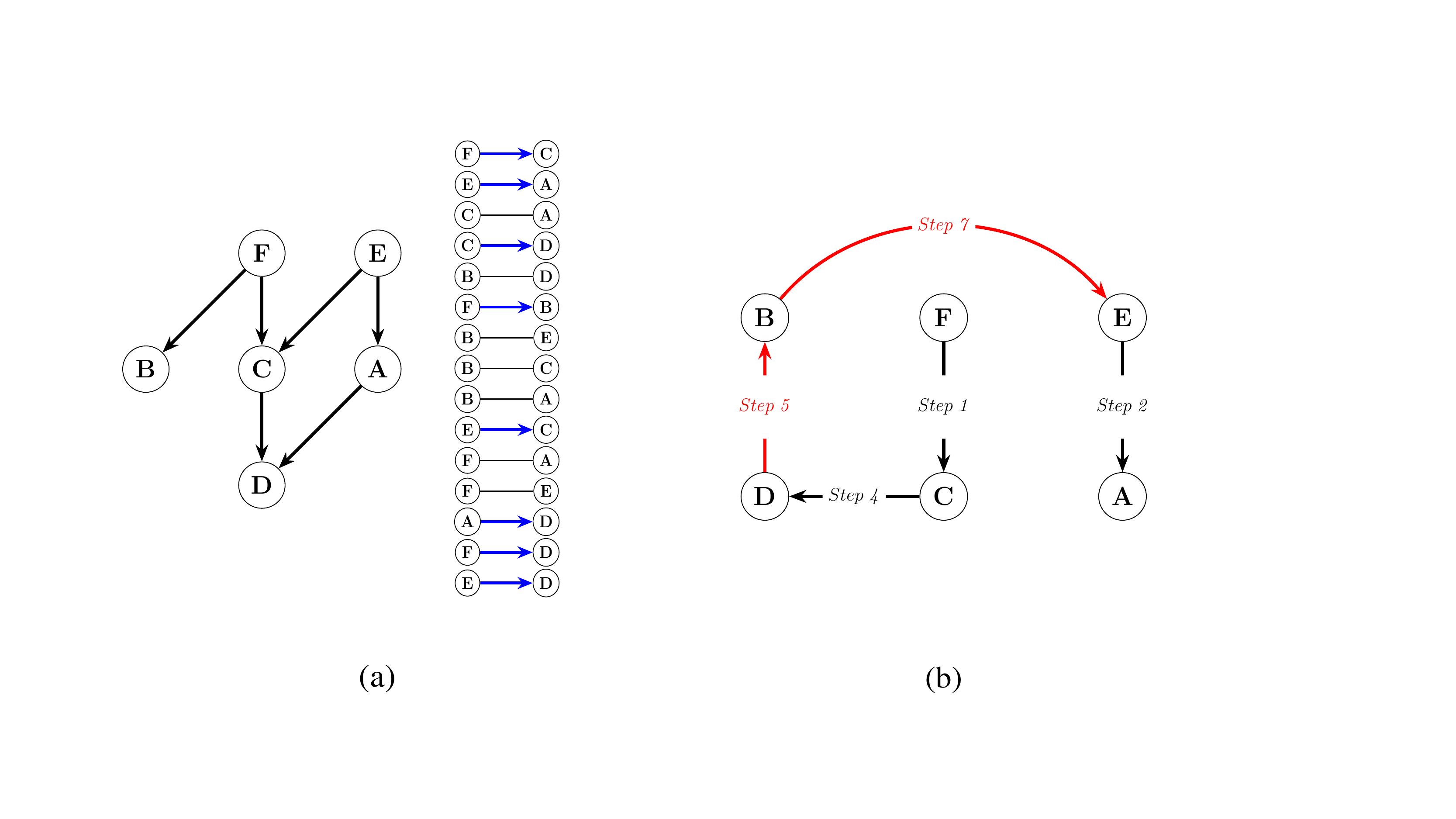}
\caption{\label{fig:frog1}An illustration of the Hie--TAN algorithm given a pre-defined DAG and a set of sorted edges.}
\end{figure}

Then Hie--MST performs dependency propagation (line 8) by considering all edges in both sets $\mathcal{D}\mathpzc{E}$ and $\mathcal{U}\mathpzc{E}$, in order to assign possible directions to the undirected edges in the set $\mathcal{U}\mathpzc{E}$. In lines 11 to 25, if those two vertices do not contain a pre-defined hierarchical dependency (indicating that edge $e(X_i,$ $X_j)$ is undirected), Hie--MST will check whether both vertices already have parents in the set $\mathcal{D}\mathpzc{E}$ in line 11. If so, the edge $e(X_i,$ $X_j)$ will not be added into  $\mathcal{UD}$, because either direction of assignment would lead to a violation of the single-parent constraint. \textit{Vice versa}, if both vertices $X_i$ and $X_j$ do not simultaneously have parents in $\mathcal{D}\mathpzc{E}$, the edge will be added into $\mathcal{U}\mathpzc{E}$, without the process of dependency propagation (line 13), because the direction of this undirected edge cannot be inferred by any other connected directed edges. 

However, if one of the vertices (either $X_i$ or $X_j$) has a parent in $\mathcal{D}\mathpzc{E}$, the undirected edge $e(X_i,$ $X_j)$ can be assigned a direction in order to satisfy the single-parent constraint. If $X_i$ has a parent in $\mathcal{D}\mathpzc{E}$, then it will be assigned as the parent of $X_j$, \textit{vice versa} for the case when $X_j$ has a parent in $\mathcal{D}\mathpzc{E}$. Then the newly directed edge $e(X_i,$ $X_j)$ will be added into $\mathcal{D}\mathpzc{E}$. After that, the process of dependency propagation will be conducted in order to assign any possible directions for those undirected edges in $\mathcal{U}\mathpzc{E}$ (lines 15 to 23).

Finally, after processing all edges by running lines 3 to 28, it is possible that there still exist some edges in $\mathcal{U}\mathpzc{E}$. As discussed in Section 2.2 and shown in Figure 1.e, some undirected edges will be added into $\mathcal{U}\mathpzc{E}$, but they cannot be assigned a direction based on the process of dependency propagation. It means that any assigned direction for those remaining undirected edges do not violate the single-parent constraint. Therefore, in lines 29 to 33, the directions for those remaining undirected edges will be randomly assigned.

To explain how Algorithms 1 and 2 work, we use the example DAG shown in Figure 2(a) (and also in Figure 1(a)), and the list of sorted edges shown in Figure 2(a). After the initialisation stage of Algorithm 1, the 1$^{st}$ step of Hie--MST is to process the edge F $\rightarrow$ C, which will be added into the set $\mathcal{D}\mathpzc{E}$ (line 7 in Algorithm 2), since it is the first edge being processed. Then in step 2, edge E $\rightarrow$ A will be processed. It will also be added into the set $\mathcal{D}\mathpzc{E}$, since it will not create a cycle and also will not lead to the violation of the single-parent constraint for all vertices in the set $\mathcal{D}\mathpzc{E}$. 

The 3$^{rd}$ step is to process the edge C -- A, which will not be added into the sets $\mathcal{U}\mathpzc{E}$ or $\mathcal{D}\mathpzc{E}$, because both vertices C and A already have parents in $\mathcal{D}\mathpzc{E}$, i.e. vertices F and E, respectively. It means that either assigning C as the parent of A or assigning A as the parent of C would violate the single-parent constraint. Then in step 4 the edge C $\rightarrow$ D will be processed and added into the set $\mathcal{D}\mathpzc{E}$ (line 7 in Algorithm 2), since adding it will not violate the single-parent constraint and will not create a cycle. 

In step 5, the edge B -- D will be processed. Although both vertices do not have a pre-defined hierarchical dependency, this edge will be added into $\mathcal{D}\mathpzc{E}$, after assigning D as the parent of B by conducting the process of dependency propagation, because vertex D is the child of vertex C in $\mathcal{D}\mathpzc{E}$. 

In step 6, when processing edge F $\rightarrow$ B, it will not be added into the set $\mathcal{D}\mathpzc{E}$, due to violation of the single-parent constraint violation for vertex B, and it would also lead to a cycle. In step 7, When processing edge B -- E, it will be also added into the set $\mathcal{D}\mathpzc{E}$ and the vertex B will be assigned as the parent of vertex E after conducting the process of dependency propagation. 

After processing all remaining edges, no edge is added into the sets $\mathcal{D}\mathpzc{E}$ or $\mathcal{U}\mathpzc{E}$, because adding those edges would lead to a cycle. Finally, the learned Hie--MST is structured as F $\rightarrow$ C $\rightarrow$ D $\rightarrow$ B $\rightarrow$ E $\rightarrow$ A, as shown in Figure 2(b).

\subsection{Hierarchical Redundancy Removed and Hierarchical Dependency Constrained Tree Augmented Na\"{i}ve Bayes}

We further propose a new type of Hie--TAN classifier, namely Hierarchical redundancy removed and hierarchical dependency constrained Tree Augmented Na\"{i}ve Bayes (Hie--TAN--Lite), which removes hierarchical redundancy between features when learning the Hie--TAN tree structure. In general, Hie--TAN--Lite removes all features that are considered as hierarchically redundant to any existing features in the tree, whilst also exploits the pre-define hierarchical dependency to determine the directions of edges in the learned tree. The pseudocode of Hie--TAN--Lite is described in Algorithms 3, 4 and 5.

\begin{figure}[!b]
\includegraphics[width=0.48\textwidth]{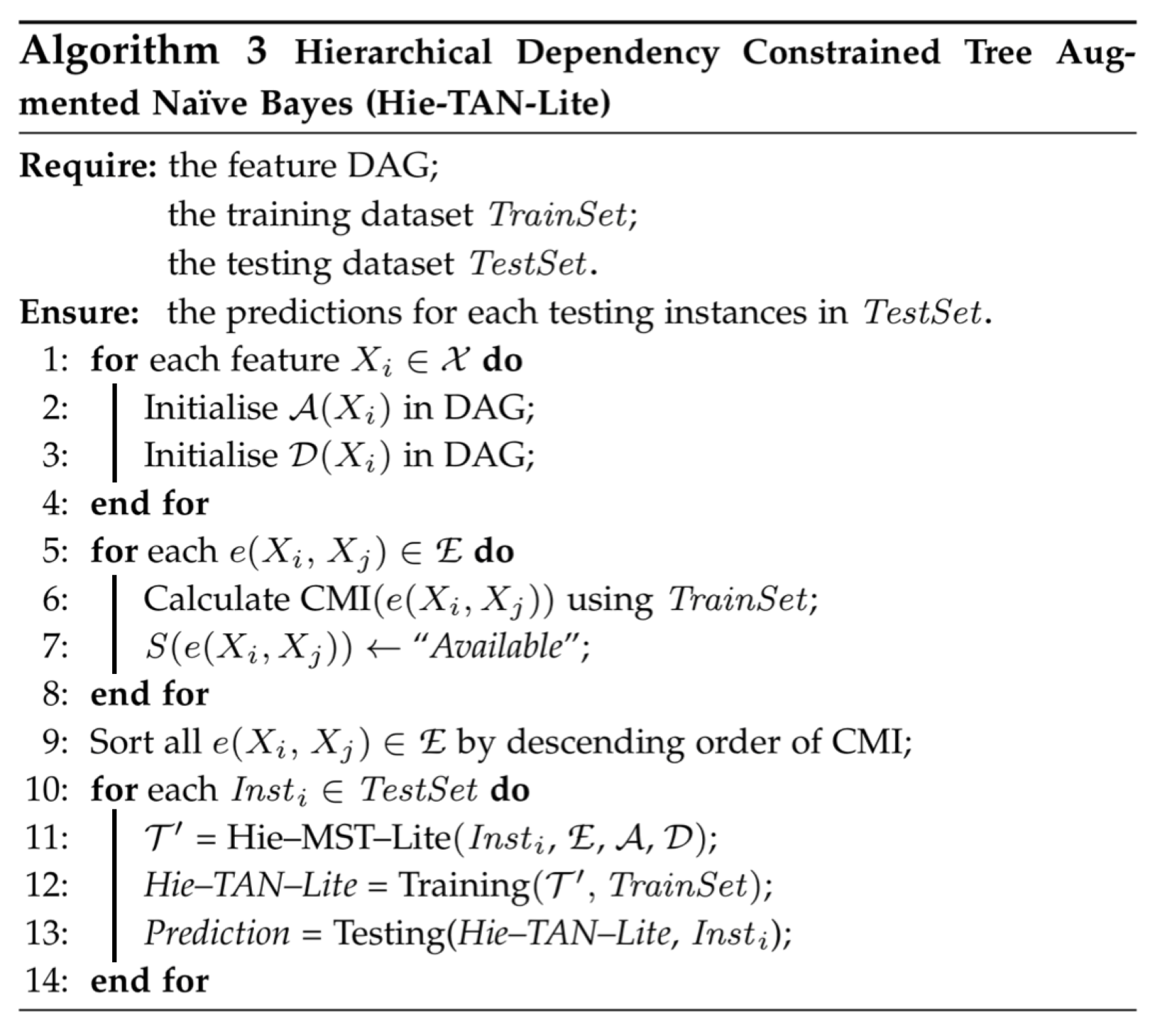}
\end{figure}

\begin{figure}[!t]
\includegraphics[width=0.48\textwidth]{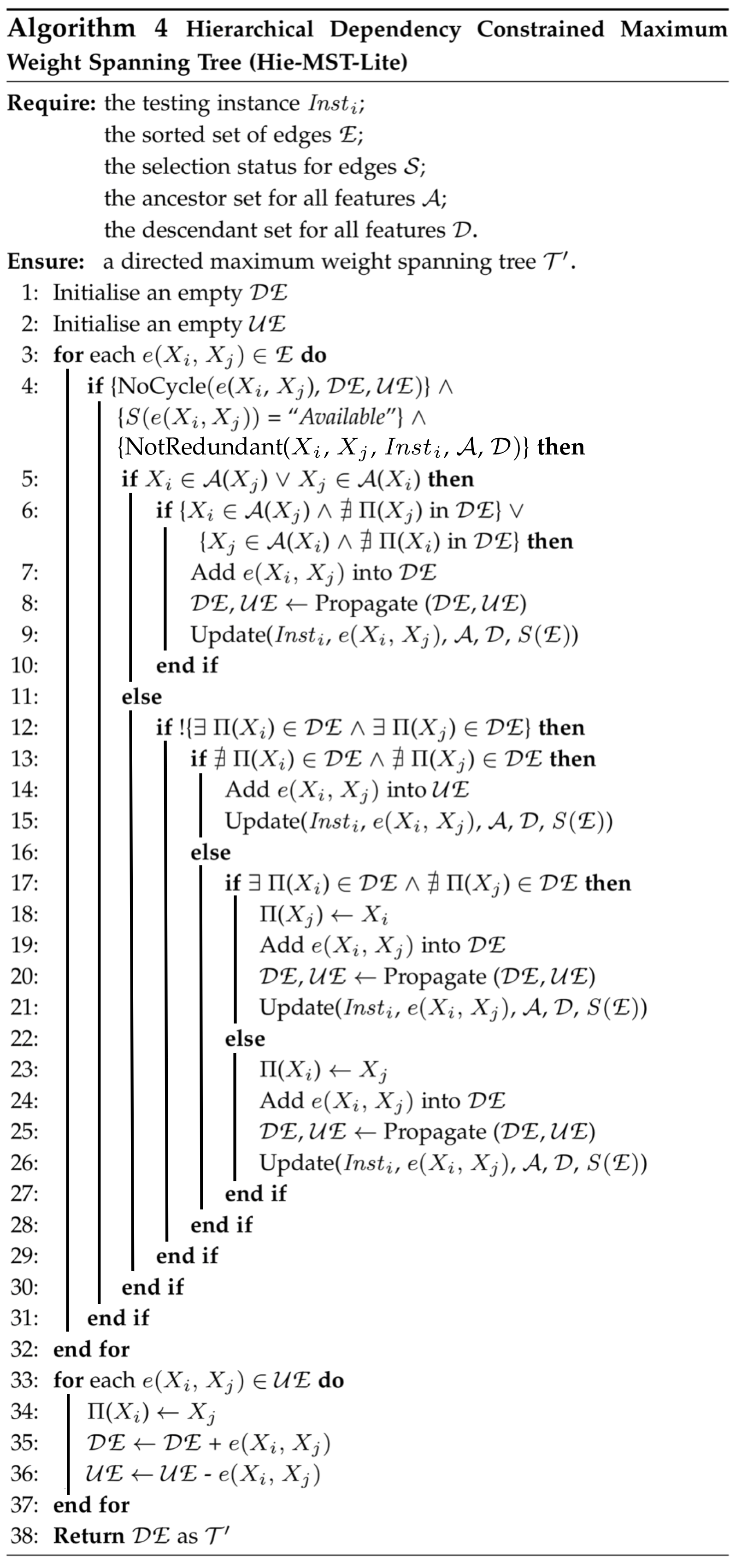}
\end{figure}

In Algorithm 3, analogously to the initialisation of Hie--TAN (Algorithm 1), Hie--TAN--Lite firstly initialises all variables that are needed for learning the Hie--TAN--Lite tree. The second stage is to learn the Hie--TAN--Lite classifier. Note that, as the hierarchical redundancy discussed in Section 2.3 occurs when considering individual testing instances, the Hie--TAN--Lite tree learning process follows the lazy learning paradigm. Hence, in lines 10 to 14, each testing instance is associated with its own separate learning phase, as follows.
For each testing instance, the $\textit{Hie--MST--Lite}$ procedure (shown in Algorithm 4) will learn a specific Hie--MST--Lite tree $\mathcal{T}'$ for that instance (line 11). Then tree $\mathcal{T}'$ is further used for training the instance-specific classifier (line 12) and for predicting the class label of that testing instance (line 13).

Algorithm 4 shows the pseudocode of the procedure Hie--MST--Lite. It firstly initialises an empty set of directed edges ($\mathcal{D}\mathpzc{E}$) and another empty set of undirected edges ($\mathcal{U}\mathpzc{E}$). Then it processes all individual edges in the sorted $\mathpzc{E}$ in lines 3 to 32. For a given edge $e(X_i,$ $X_j)$, line 4 checks whether adding this edge will lead to a cycle by considering all existing edges in both sets $\mathcal{D}\mathpzc{E}$ and $\mathcal{U}\mathpzc{E}$. If adding that edge $e(X_i,$ $X_j)$ will not lead to a cycle, Hie--MST--Lite checks whether the status of that edge is available, then it checks whether the pair of vertices in that edge are hierarchically redundant given their values in that specific testing instance. After checking the criteria of hierarchical redundancy, in lines 5 to 30, Hie--MST--Lite continues to process that edge by considering the pre-defined hierarchical dependency constraint as analogous to Hie--MST. However, note that, in order to remove the hierarchical redundancy, Hie--MST--Lite also adopts the \textit{RemoveRedundancy} procedure after adding any edges into $\mathcal{T}'$, as shown in lines 9, 15, 21 and 26.

Algorithm 5 shows the pseudocode of the procedure \textit{RemoveRedundancy}. In lines 1 to 9, \textit{RemoveRedundancy} firstly checks whether any of the two vertices (features) in the current edge are hierarchically redundant with respect to their  ancestor and descendant sets (i.e. if any of those two vertices have the same feature value $\mathcal{V}$ as some of its ancestor or descendant vertices) in the current testing instance $\mathit{Inst}_{i}$ (lines 2 and 3). Then, in order to remove those hierarchically redundant vertices, the status of all edges consisting of those redundant features will be assigned as \textit{Unavailable} (lines 4 to 6). Finally, \textit{RemoveRedundancy} returns the updated set of status for all vertices.

\begin{figure}[t]
\includegraphics[width=0.48\textwidth]{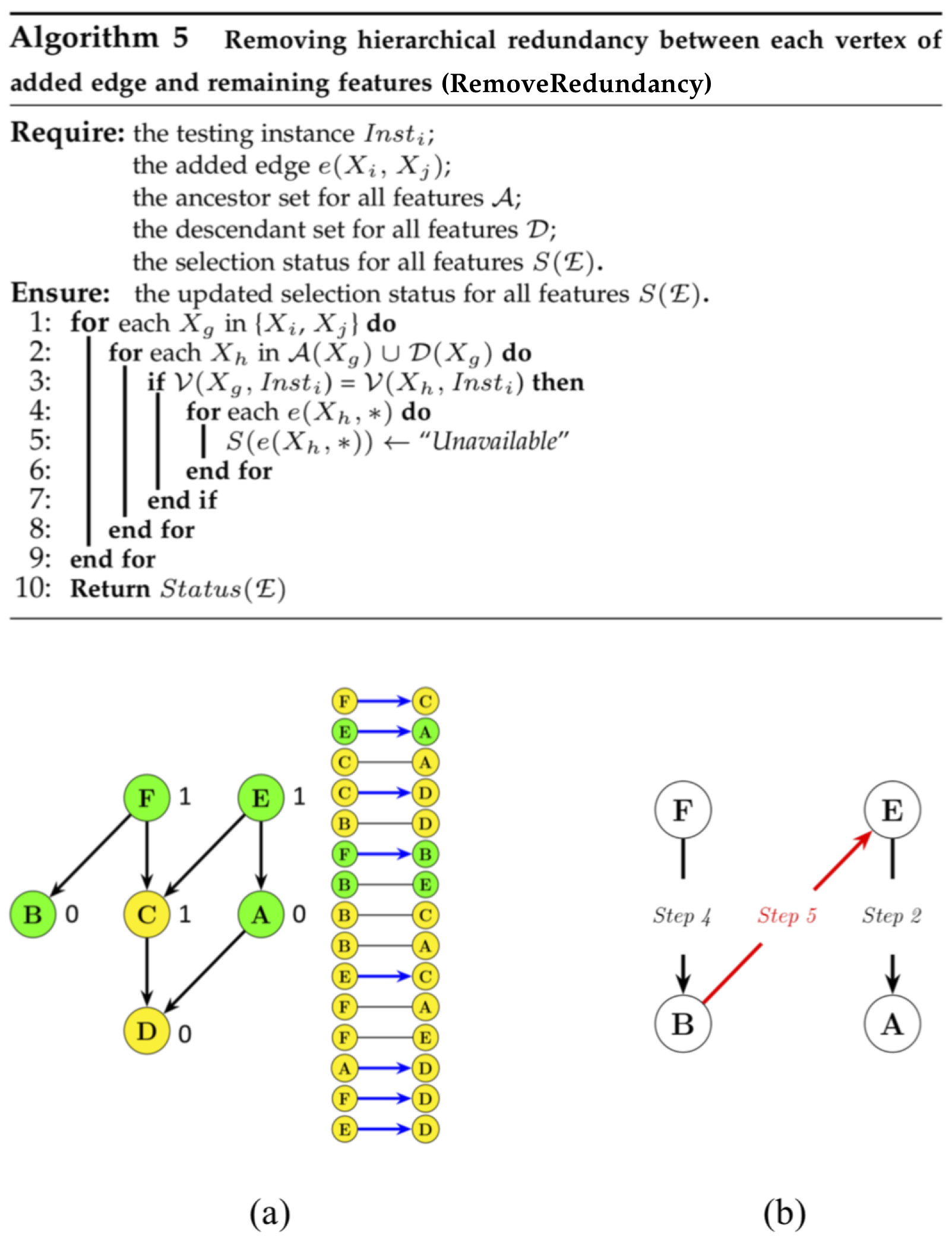}
\caption{\label{fig:frog1}An illustration of the Hie--TAN--Lite algorithm given a pre-defined DAG and sorted edges.}
\vspace{-0.6cm}
\end{figure}

To explain how Algorithms 3, 4 and 5 work, we use the example DAG and the list of sorted edges shown in Figure 3(a) (and also in Figure 2(a)). After the initialisation stage of Algorithm 3, Hie--TAN--Lite starts to learn one specific tree for each individual testing instance. The first step of Hie--MST--Lite is to process the edge F $\rightarrow$ C. Unlike Hie--MST, Hie--MST--Lite will not add that edge into the set $\mathcal{D}\mathpzc{E}$, since vertex F is the parent of vertex C, and both vertices have the same value \textit{1}. 

In step 2, the directed edge E $\rightarrow$ A will be processed. It will be added into the set $\mathcal{D}\mathpzc{E}$, since both vertices E and A have different values, although E is the parent of A, meaning there is no hierarchical redundancy between these vertices. Also, adding edge E $\rightarrow$ A will not create a cycle and will not lead to the violation of the single-parent constraint for vertices in the set $\mathcal{D}\mathpzc{E}$. After adding the directed edge E $\rightarrow$ A, the third step is to remove all other edges from the candidate edge set, in order to avoid any potential hierarchical redundancy between vertices. Because the value of vertex E equals to \textit{1}, vertex C is considered as a hierarchically redundant vertex, since it is the child of vertex E and also has the value of \textit{1} in the current testing instance. Hence, all edges that include vertex C will be removed from the candidate edge set, as shown in Algorithm 5, lines 2 to 8, where the status of the corresponding candidate edges will be assigned as $\textit{Unavailable}$. Analogously, all other vertices that are hierarchically redundant to vertex A will be removed, such as vertex D, which has the same value to vertex A, and is the child of vertex A. After processing the third step, edges C -- A, C -- D, B -- D, B -- C, E -- C, A -- D, F -- D, and E -- D are removed, as shown in yellow in Figure 3(a).

In step 4, the directed edge F $\rightarrow$ B will be processed. This edge will be added into $\mathcal{D}\mathpzc{E}$, as it satisfies all criteria shown in lines 4 to 6 of Algorithm 4. As vertex C has already been removed, there is no other relevant vertex to be removed. In step 5, the undirected edge B -- E will be processed. It will be added into $\mathcal{D}\mathpzc{E}$, since it satisfies all criteria, and the direction can be determined by considering the single-parent constraint, i.e. vertex B has already had a parent F, then vertex B should be defined as the parent of vertex E. As there is no more candidate edges to be processed, the remaining features are F, B, E and A, as shown in Figure 3(a), where the nodes in yellow denote the removed features and corresponding edges, and the nodes in green mean the remaining ones. Figure 3(b) shows the final Hie--MST--Lite tree, structured as F $\rightarrow$ B $\rightarrow$ E $\rightarrow$ A.

\section{Computational experiments}
\subsection{Datasets and Experimental Methodology}
We evaluate the predictive performance of the proposed Hie--TAN and Hie--TAN--Lite algorithms by using 28 bioinformatics datasets of ageing-related genes~\cite{WanICML,PGM2020}. In these datasets, the instances to be classified are genes, the binary features denote the presence or absence of GO term annotations for the genes, and the binary class variable indicates whether a gene has a pro- or anti-longevity effect.

\begin{figure}[t]
\includegraphics[width=0.4\textwidth]{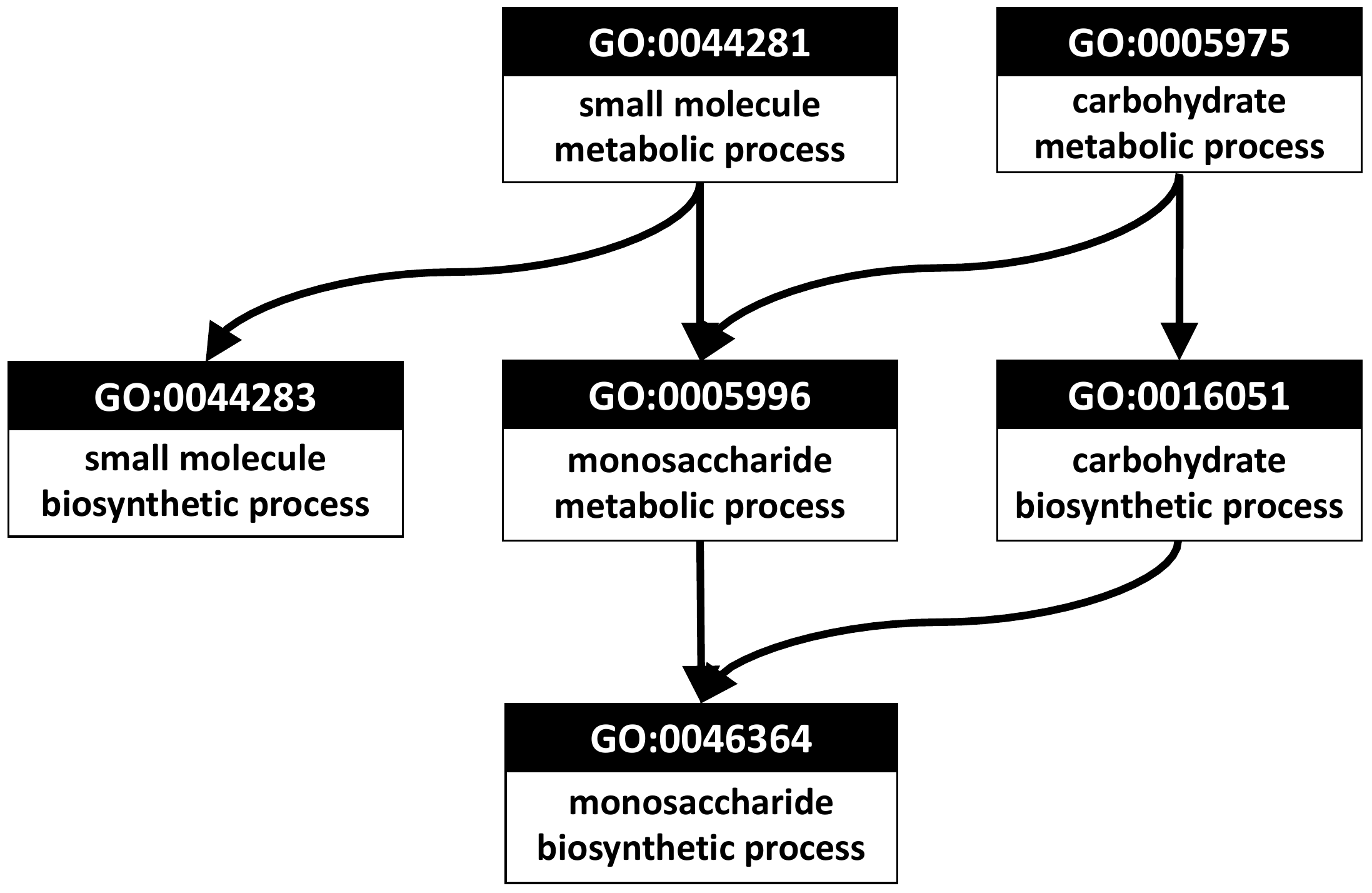}
\caption{\label{fig:frog1}An example Gene Ontology hierarchy.}
\end{figure}

Figure 4 shows 6 example GO terms and their hierarchically relationships, e.g. GO:0044281 (\textit{small molecule metabolic process}) is the parent of GO:0044283 (\textit{small molecule biosynthetic process}) and GO:0005996 (\textit{monosaccharide metabolic process}), which is also the parent of GO:0046364 (\textit{monosaccharide biosynthetic process}). This type of hierarchy therefore is a Directed Acyclic Graph (DAG), as shown in Figure 1(a), where those 6 GO terms were represented by 6 arbitrary letters: A, B, C, D, E and F.

In this work, we use GO terms as predictive binary features to describe genes. For each gene (instance), the GO term's feature values $\textit{1}$ and $\textit{0}$ denote that gene is or is not annotated with that GO term, respectively. According to the pre-defined hierarchical dependency between GO terms in the GO DAG, the value $\textit{1}$ of each GO term feature is propagated to all its corresponding ancestor GO term features (i.e. if a gene is annotated with a GO term, then that gene is always annotated with all its ancestors GO terms), which leads to a sparse matrix of binary feature values. Figure 5 shows an example dataset including six features whose pre-defined hierarchical dependencies are discussed in Figure 1(a). As feature D is the descendant of other four features (F, C, E and A), as shown for \textit{Instance\_1} in Figure 5, if the value of feature D is \textit{1}, then the values of all those four features will be \textit{1}. \textit{Vice versa}, as feature E is the ancestor of features C, A and D, if the value of feature E is \textit{0}, then the values of all those three features will be \textit{0}, as shown for \textit{Instance\_2} in Figure 5.

\begin{figure}[!t]
    \centering
        \includegraphics[width=6.5cm]{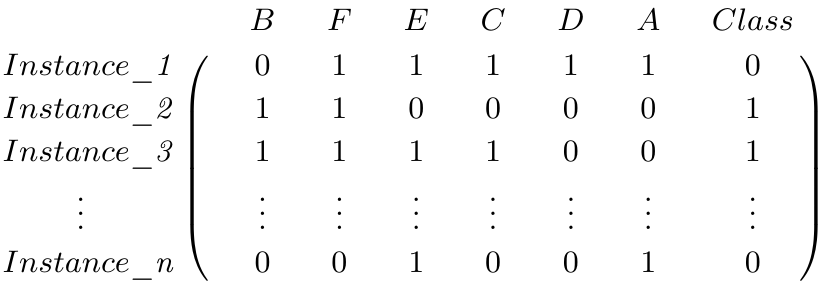}
        \caption{An example matrix of dataset including six features and generated according to their pre-defined hierarchical dependencies information.}
\vspace{-0.47cm}
\end{figure}

We compare the predictive accuracy of the proposed Hie--TAN and Hie--TAN--Lite with four other methods: Hie--AODE--Lite~\cite{PGM2020}, HRE--TAN~\cite{WanICML}, TAN~\cite{Friedman1997} and proximalGraph~\cite{GraphLASSO1}. TAN is used as a natural baseline method which ignores the pre-defined feature DAG, whilst the other methods were all designed specifically for coping with a pre-defined feature DAG. Hie--AODE--Lite is an ensemble of one-dependence estimators (ODEs). Each ODE consists of one parent node with outward edges pointing to all other features except its ancestors, in order to satisfy the pre-defined hierarchical dependency constraint. HRE--TAN exploits pre-defined hierarchical dependencies to learn a tree-like structure of features with no hierarchical redundancy, i.e. after adding individual candidate edges into the tree, the ancestor or descendant features of any vertices of that edge will be removed depending on their values in specific testing instances. ProximalGraph exploits a type of sparsity-inducing regularisation function and the proximal operator (an extension of gradient-based optimisation method) to learn a set of non-zero coefficients for a linear model to cope with data where features are organised into pre-defined hierarchical dependencies. We evaluated the algorithms using 10-fold cross-validation and the Geometric Mean (GMean) of Sensitivity and Specificity as the predictive accuracy evaluation metric -- i.e. the square root of the product of Sensitivity and Specificity.

\begin{figure*}[!t]
\includegraphics[width=1\textwidth]{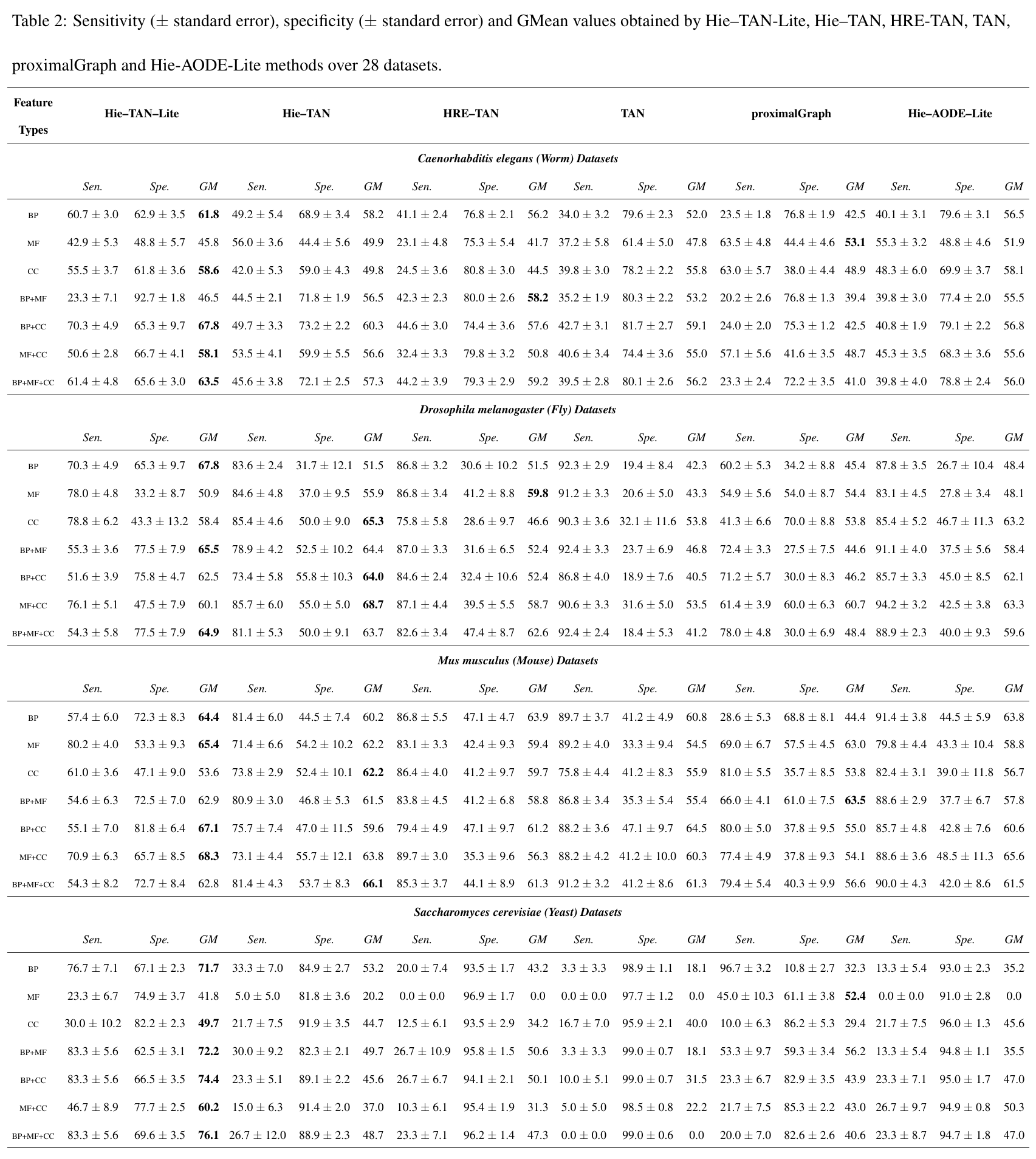}
\end{figure*}

\subsection{Experimental results}
Table 1 shows the experimental results obtained by the 6 classification methods. In each row of this table (i.e., for each dataset), the highest GMean value for that dataset is shown in boldface font. We also compute the average rank for each method, based on the GMean measure, as follows. 

First, for each dataset, the method with the highest GMean is assigned rank 1, whilst the method with the lowest GMean is assigned rank 6. Tied rank numbers are divided among the tied methods -- e.g. if two methods are tied as the best methods, each method is assigned rank 1.5. Then, we compute each method's average rank over the 28 datasets. Note that, the lower the rank, the better the method.

In general, Hie--TAN--Lite was the best-performing method due to its best average rank of 1.89. It also obtained the highest GMean values in 18 out of the 28 datasets. The second best-performing method was Hie--TAN, which obtained the average rank of 2.59 and the highest GMean values in 5 out of the 28 datasets. The Hie--AODE--Lite algorithm obtained an average rank of 3.43, which is still substantially better than average ranks obtained by HRE--TAN, proximalGraph and the conventional TAN methods. 

We used the well-known Friedman test with Holm's \textit{post-hoc} multiple-hypothesis correction~\cite{Demsar2006} to compare the average ranks of GMean values obtained by the top-performing method (Hie--TAN--Lite) against each other method. The statistical test results confirm that the proposed Hie--TAN--Lite significantly outperformed nearly all other methods (i.e. Hie--AODE--Lite, HRE--TAN, proximalGraph and TAN, with Holm p-values 1.04 E-03, 1.36 E-04, 7.20 E-08 and 1.43 E-09, respectively), except Hie--TAN (also proposed in this work).

\begin{figure}[!t]
\centering
\resizebox{8.0cm}{!}{
\renewcommand{\arraystretch}{2.3}
\begin{tabular}{ccccc}
\multicolumn{4}{c}{\multirow{2}{*}{\LARGE Table 3: Results of Friedman test and Holm \textit{post-hoc} correction.}}\\
\multicolumn{4}{c}{\multirow{2}{*}{}}\\\hline
\LARGE Method & \LARGE \# Wins & \LARGE Average Ranking & \LARGE Adjusted $\alpha$ & \LARGE P-value \\\hline
\LARGE\bf Hie--TAN--Lite &\LARGE 18 & \LARGE 1.89 & \LARGE N/A & \LARGE N/A \\ 
\LARGE\bf Hie--TAN &\LARGE 5 & \LARGE 2.59 & \LARGE 5.00 E-02 & \LARGE \LARGE 8.08 E-02 \\ 
\LARGE Hie--AODE--Lite &\LARGE 0& \LARGE 3.43 & \LARGE 2.50 E-02 & \LARGE\underline{1.04 E-03}\\
\LARGE HRE--TAN &\LARGE 2&\LARGE 3.71 & \LARGE 1.67 E-02 & \LARGE\underline{1.36 E-04} \\
\LARGE proximalGraph &\LARGE 3& \LARGE 4.52 & \LARGE 1.25 E-02 & \LARGE\underline{7.20 E-08} \\
\LARGE TAN &\LARGE 0& \LARGE 4.86 & \LARGE 1.00 E-02 &\LARGE \underline{1.43 E-09} \\\hline
\end{tabular}
}
\end{figure}

\subsection{Identifying the GO Terms (Features) Most Often Used for Classification}
As the proposed Hie--TAN--Lite method outperformed all other methods in general, we report the GO terms most frequently selected by Hie--TAN--Lite in the BP datasets for each of the 4 model organisms. Since Hie--TAN--Lite removes edges with hierarchically redundant features, we report two types of ranking criteria: \textit{Freq. of Selection} and \textit{Freq. in Edges}. The former means the number of testing instances for which the GO term was selected to be included in the Hie--TAN--Lite tree, whilst the latter means the number of edges containing the GO term in the Hie--TAN--Lite trees, over all testing instances.

\indent Table S1 shows the top-ranked GO terms for the four model organisms' datasets. In general, \textit{reproduction} (GO:0000003) process-related terms were highly relevant to ageing: \textit{single organism reproductive process} (GO:0044702) for all four organisms' datasets; \textit{developmental process involved in reproduction} (GO:0003006) for the fly and yeast datasets; and \textit{cellular process involved in reproduction} (GO:0048610) for the yeast dataset. It has been found that removing germ cells of worms extended their lifespan by about 60\% \citep{KenyonRe3}, and transplanting young mice's ovaries into old recipients also extend their lifespan \citep{KenyonRe1,KenyonRe2}. These findings are related to biological pathways that regulate the reproduction and ageing processes; e.g. the insulin/IGF-1 signalling \citep{KenyonRe2} regulates the activity of DAF-16/FOXO -- a key ageing-related transcription factor \citep{KenyonMeta1,KenyonNutri1,Vellai2003,Berdichevsky2006}.

Several metabolism-related GO terms were also among the top-ranked GO terms in Table S1: \textit{heterocycle metabolic process} (GO:0046483) in the worm, mouse and yeast datasets; \textit{organic substance metabolic process} (GO:0071704) in the worm dataset; \textit{organic substance transport} (GO:0071702), \textit{regulation of primary metabolic process} (GO:0080090), and \textit{cellular aromatic compound metabolic process} (GO:0006725) in the yeast dataset. These findings are consistent with research showing that ageing is closely related to nutrient metabolism pathways, like the target of rapamycin (TOR) signalling pathway -- inhibiting the TOR pathway extends multiple species' lifespan \citep{Kaeberlein2005,Kapahi2004}.

\section{Conclusions}
We proposed two novel tree augmented na\"{i}ve Bayes classification algorithms that exploit pre-defined hierarchical dependencies among features. The results showed that the two proposed methods not only successfully improved the accuracy of the conventional TAN method, but also outperformed other methods that also exploit hierarchical feature dependencies. An interesting future research direction would be to propose other Bayesian network classification algorithms for exploiting the hierarchical feature dependencies.
\vspace{-0.3cm}

\section*{Funding}
This work is supported by the Birkbeck School Research Grant.

\nocite{langley00}

\bibliography{example_paper.bbl}

\providecommand{\noopsort}[1]{}
\begin{thebibliography}{24}
\providecommand{\natexlab}[1]{#1}
\providecommand{\url}[1]{\texttt{#1}}
\expandafter\ifx\csname urlstyle\endcsname\relax
  \providecommand{\doi}[1]{doi: #1}\else
  \providecommand{\doi}{doi: \begingroup \urlstyle{rm}\Url}\fi

\bibitem[Berdichevsky et~al.(2006)Berdichevsky, Viswanathan, Horvitz, and
  Guarente]{Berdichevsky2006}
Berdichevsky, A., Viswanathan, M., Horvitz, H., and Guarente, L.
\newblock C. elegans sir-2.1 interacts with 14-3-3 proteins to activate daf-16
  and extend life span.
\newblock \emph{Cell}, 125:\penalty0 1165--1177, 2006.

\bibitem[da~Silva et~al.(2018)da~Silva, Plastino, and Freitas]{SDM18}
da~Silva, P.~N., Plastino, A., and Freitas, A.~A.
\newblock A novel genetic algorithm for feature selection in hierarchical
  feature spaces.
\newblock In \emph{Proceedings of the 2018 SIAM International Conference on
  Data Mining}, pp.\  738--746, San Diego, USA, 2018.

\bibitem[da~Silva et~al.(2020)da~Silva, Plastino, and Freitas]{Pablo2020}
da~Silva, P.~N., Plastino, A., and Freitas, A.~A.
\newblock Prioritizing positive feature values: a new hierarchical feature
  selection method.
\newblock \emph{Applied Intelligence}, 50:\penalty0 4412–4433, 2020.

\bibitem[{de Magalh\~{a}es} et~al.(2009){de Magalh\~{a}es}, Budovsky, Lehmann,
  Costa, Li, Fraifeld, and Church]{DeMagalhaes2009}
{de Magalh\~{a}es}, J.~P., Budovsky, A., Lehmann, G., Costa, J., Li, Y.,
  Fraifeld, V., and Church, G.~M.
\newblock The human ageing genomic resources: online databases and tools for
  biogerontologists.
\newblock \emph{Aging Cell}, 8\penalty0 (1):\penalty0 65--72, February 2009.

\bibitem[Dems\v{a}r(2006)]{Demsar2006}
Dems\v{a}r, J.
\newblock Statistical comparisons of classifiers over multiple data sets.
\newblock \emph{The Journal of Machine Learning Research}, 7:\penalty0 1--30,
  January 2006.

\bibitem[Friedman et~al.(1997)Friedman, Geiger, and Goldszmidt]{Friedman1997}
Friedman, N., Geiger, D., and Goldszmidt, M.
\newblock {B}ayesian network classifiers.
\newblock \emph{Machine Learning}, 29:\penalty0 131--163, November 1997.

\bibitem[Hansen et~al.(2007)Hansen, Taubert, Crawford, Libina, Lee, and
  Kenyon]{KenyonNutri1}
Hansen, M., Taubert, S., Crawford, D., Libina, N., Lee, S., and Kenyon, C.
\newblock Lifespan extension by conditions that inhibit translation in
  caenorhabditis elegans.
\newblock \emph{Aging Cell}, 6:\penalty0 95--110, 2007.

\bibitem[Hsin \& Kenyon(1999)Hsin and Kenyon]{KenyonRe3}
Hsin, H. and Kenyon, C.
\newblock Signals from the reproductive system regulate the lifespan of c.
  elegans.
\newblock \emph{Nature}, 399:\penalty0 362–366, 1999.

\bibitem[Jenatton et~al.(2011{\natexlab{a}})Jenatton, Audibert, and
  Bach]{LASSOFS3}
Jenatton, R., Audibert, J., and Bach, F.
\newblock Structured variable selection with sparsity-inducing norms.
\newblock \emph{Journal of Machine Learning Research}, 12:\penalty0 2777--2824,
  2011{\natexlab{a}}.

\bibitem[Jenatton et~al.(2011{\natexlab{b}})Jenatton, Mairal, Obozinski, and
  Bach]{GraphLASSO2}
Jenatton, R., Mairal, J., Obozinski, G., and Bach, F.
\newblock Proximal methods for hierarchical sparse coding.
\newblock \emph{Journal of Machine Learning Research}, 12:\penalty0 2297--2334,
  2011{\natexlab{b}}.

\bibitem[Kaeberlein et~al.(2005)Kaeberlein, 3rd, Steffen, Westman, Hu, Dang,
  Kerr, Kirkland, Fields, and Kennedy]{Kaeberlein2005}
Kaeberlein, M., 3rd, R.~P., Steffen, K., Westman, E., Hu, D., Dang, N., Kerr,
  E., Kirkland, K., Fields, S., and Kennedy, B.
\newblock Regulation of yeast replicative life span by tor and sch9 in response
  to nutrients.
\newblock \emph{Science}, 310:\penalty0 1193--1196, 2005.

\bibitem[Kapahi et~al.(2004)Kapahi, Zid, Harper, Koslover, Sapin, and
  Benzer]{Kapahi2004}
Kapahi, P., Zid, B., Harper, T., Koslover, D., Sapin, V., and Benzer, S.
\newblock Regulation of lifespan in drosophila by modulation of genes in the
  tor signaling pathway.
\newblock \emph{Current Biology}, 14:\penalty0 885--890, 2004.

\bibitem[Kenyon(2005)]{KenyonRe2}
Kenyon, C.
\newblock The plasticity of aging: insights from long-lived mutants.
\newblock \emph{Cell}, 120:\penalty0 449–460, 2005.

\bibitem[Kenyon(2010)]{KenyonRe1}
Kenyon, C.
\newblock A pathway that links reproductive status to lifespan in
  caenorhabditis elegans.
\newblock \emph{Annals of The New York Academy of Sciences}, 1204:\penalty0
  156--162, 2010.

\bibitem[Lee et~al.(2009)Lee, Murphy, and Kenyon]{KenyonMeta1}
Lee, S., Murphy, C., and Kenyon, C.
\newblock Glucose shortens the lifespan of c. elegans by downregulating
  daf-16/foxo activity and aquaporin gene expression.
\newblock \emph{Cell Metabolism}, 10:\penalty0 379–391, 2009.

\bibitem[Mairal \& Yu(2013)Mairal and Yu]{LASSOFS2}
Mairal, J. and Yu, B.
\newblock Supervised feature selection in graphs with path coding penalties and
  network flows.
\newblock \emph{Journal of Machine Learning Research}, 14:\penalty0 2449--2485,
  2013.

\bibitem[Mairal et~al.(2010)Mairal, Jenatton, Obozinski, and Bach]{GraphLASSO1}
Mairal, J., Jenatton, R., Obozinski, G., and Bach, F.
\newblock Network flow algorithms for structured sparsity.
\newblock In \emph{Proceedings of the 2010 Advances Neural Information
  Processing Systems}, pp.\  1558--1566, Vancouver, Canada, 2010.

\bibitem[Ristoski \& Paulheim(2014)Ristoski and Paulheim]{SHSEL}
Ristoski, P. and Paulheim, H.
\newblock Feature selection in hierarchical feature spaces.
\newblock In \emph{Proceedings of the International Conference on Discovery
  Science (DS 2014)}, pp.\  288--300, Bled, Slovenia, 2014.

\bibitem[{The Gene Ontology Consortium}(2000)]{GO2000}
{The Gene Ontology Consortium}.
\newblock Gene {O}ntology: tool for the unification of biology.
\newblock \emph{Nature Genetics}, 25\penalty0 (1):\penalty0 25--29, May 2000.

\bibitem[Vellai et~al.(2003)Vellai, Takacs-Vellai, Zhang, Kovacs, Orosz, and
  Müller]{Vellai2003}
Vellai, T., Takacs-Vellai, K., Zhang, Y., Kovacs, A., Orosz, L., and Müller,
  F.
\newblock Genetics: influence of tor kinase on lifespan in c. elegans.
\newblock \emph{Nature}, 426:\penalty0 620, 2003.

\bibitem[Wan \& Freitas{\noopsort{c}}(2020)Wan and
  Freitas{\noopsort{c}}]{PGM2020}
Wan, C. and Freitas{\noopsort{c}}, A.~A.
\newblock Hierarchical dependency constrained averaged one-dependence
  estimators classifiers for hierarchical feature spaces.
\newblock In \emph{Proceedings of the 10th International Conference on
  Probabilistic Graphical Models}, volume 138, pp.\  557--568, Aalborg,
  Denmark, 2020. Proceedings of Machine Learning Research.

\bibitem[Wan \& Freitas{\noopsort{h}}(2016)Wan and
  Freitas{\noopsort{h}}]{WanICML}
Wan, C. and Freitas{\noopsort{h}}, A.~A.
\newblock A new hierarchical redundancy eliminated tree augmented naive bayes
  classifier for coping with gene ontology-based features.
\newblock In \emph{Proceedings of the 33rd International Conference on Machine
  Learning (ICML 2016) Workshop on Computational Biology}, New York, USA, 5
  pp., 2016.

\bibitem[Wan \& Freitas{\noopsort{k}}(2015)Wan and
  Freitas{\noopsort{k}}]{WanACMBCB2015}
Wan, C. and Freitas{\noopsort{k}}, A.~A.
\newblock Two methods for constructing a gene ontology-based feature selection
  network for a {B}ayesian network classifier and applications to datasets of
  aging-related genes.
\newblock In \emph{Proceedings of the Sixth ACM Conference on Bioinformatics,
  Computational Biology and Health Informatics (ACM-BCB 2015)}, pp.\  27--36,
  Atlanta, USA, 2015.

\bibitem[Wan et~al.(2015)Wan, Freitas, and {de Magalh\~{a}es}]{Wan2014}
Wan, C., Freitas, A.~A., and {de Magalh\~{a}es}, J.~P.
\newblock Predicting the pro-longevity or anti-longevity effect of model
  organism genes with new hierarchical feature selection methods.
\newblock \emph{IEEE/ACM Transactions on Computational Biology and
  Bioinformatics}, 12\penalty0 (2):\penalty0 262--275, March 2015.

\end{thebibliography}
\bibliographystyle{icml2020}

\section*{Supplementary Material}
\begin{figure*}[!b]
\centering
\vspace{-0.5cm}
\includegraphics[width=1\textwidth]{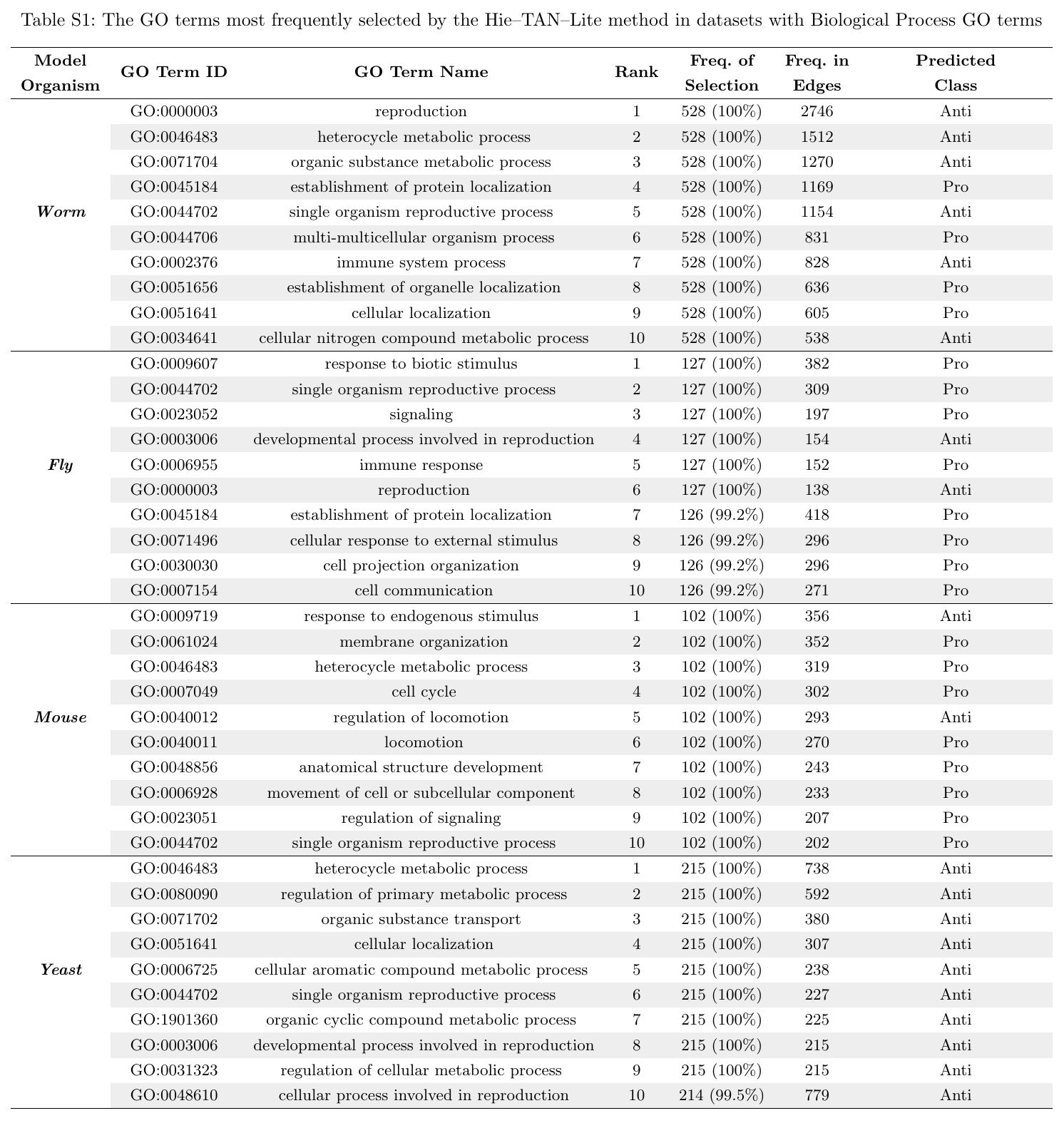}
\vspace{-0.3cm}
\end{figure*}


\end{document}